\documentclass[11pt,letterpaper]{article}
\usepackage{graphicx}
\usepackage{fancyhdr}
\usepackage{lastpage}
\usepackage{color}
\usepackage{wrapfig}
\usepackage{hyperref}
\usepackage{ownstyles}

\graphicspath{{../figures/}}

\title{MAV Stabilization using Machine Learning and \\Onboard Sensors\\{\small \sc CS6780 Final Report}}
\author{Jason Yosinski and Cooper Bills \\
\code{yosinski@cs.cornell.edu, csb88@cornell.edu}}
\date{December 17, 2010}

\begin{document}
\maketitle

\begin{abstract}
In many situations, Miniature Aerial Vehicles (MAVs) are limited to
using only on-board sensors for navigation. This limits the data
available to algorithms used for stabilization and localization, and
current control methods are often insufficient to allow reliable
hovering in place or trajectory following.  In this research, we
explore using machine learning to predict the drift (flight path
errors) of an MAV while executing a desired flight path.  This
predicted drift will allow the MAV to adjust it's flightpath to
maintain a desired course.

\end{abstract}


\section{Introduction}

Past automation work with miniature aerial vehicles (MAVs) at Cornell
has produced interesting results \cite{Bills11} and presented
additional challenges.  During past projects, results have often been
limited not by insufficiencies in planning algorithms, but by
navigation errors stemming from inadequate control in the face of
realistic, breezy operating environments.  In many cases the MAVs will
simply drift off the desired path (\figref{path.jpg}). Thus, this
project focuses on refining the basic motion of the same
platform, and in particular, minimizing its drift.

Our work focuses on reduction of low frequency drift in gps-denied
environments.  Similar work has been done, some using neural networks
\cite{Nicol08} or using adaptive-fuzzy control methods \cite{Coza06}
to stabilize a quadrotor.  Though this research has produced promising
results, these methods were demonstrated only in simulation, not via
live testing.
\clearpage

\figvarp{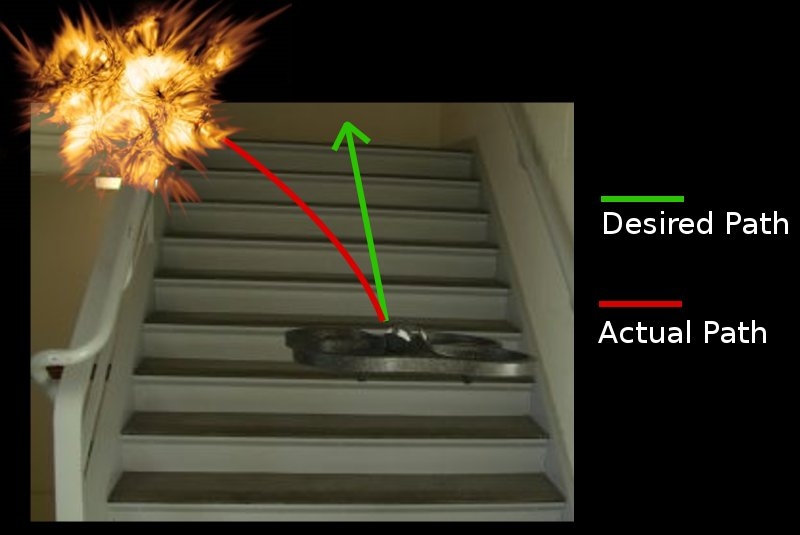}{.6}{Desired path vs. actual path due to drift.}{}
\section{Platform}

\begin{wrapfigure}[9]{r}{2.1in}
  \vspace{-20pt}
  \begin{center}
    \includegraphics[width=2.1in]{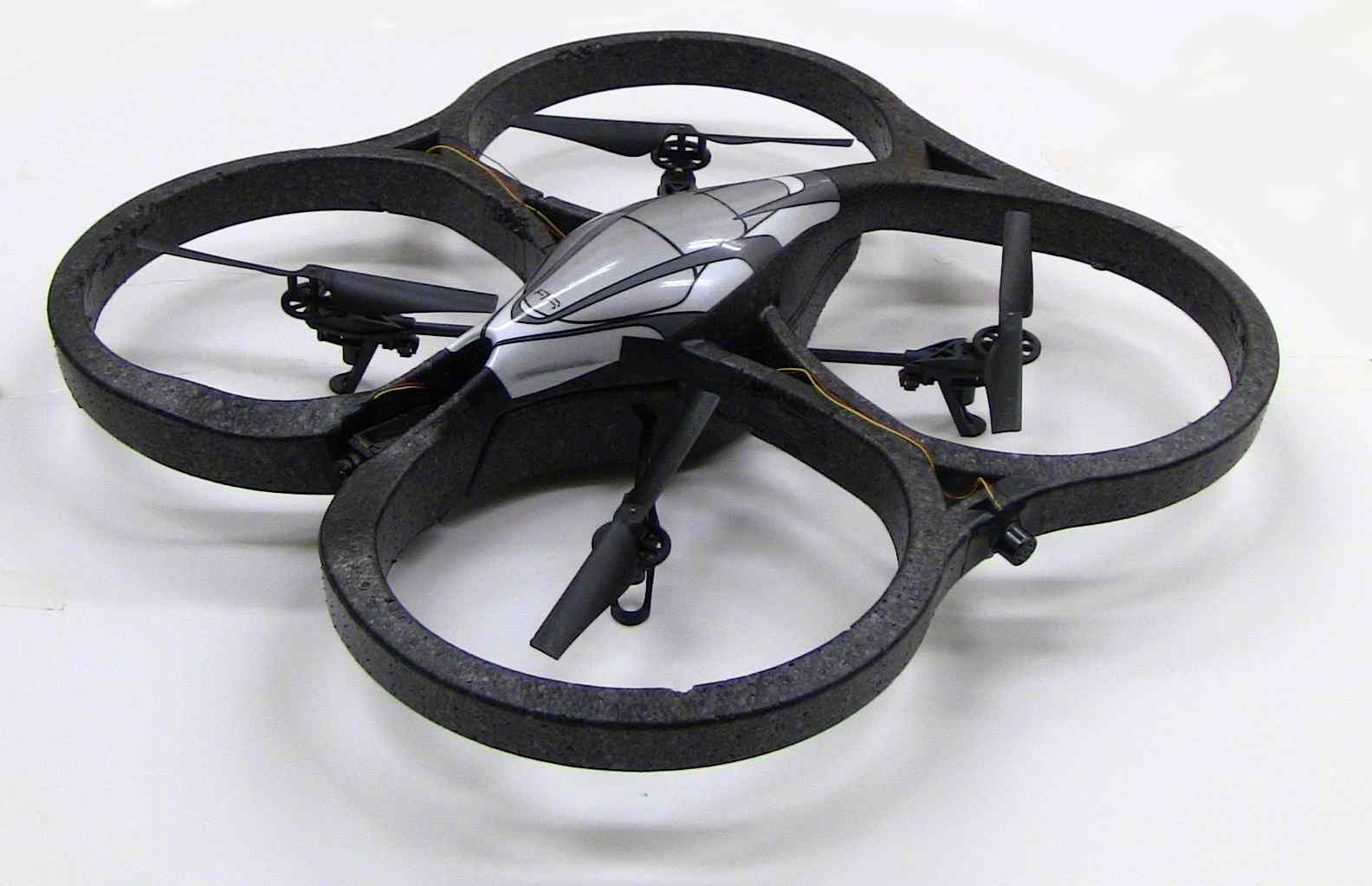}
  \end{center}
  \vspace{-20pt}
  \caption{Our Parrot AR.Drone Platform}
\figlabel{quadrotor}
\end{wrapfigure}

We are using the Parrot AR.Drone quadrotor (\figref{quadrotor}), a toy
recently available to the general public.  Commands and images are exchanged
via a WiFi ad-hoc connection between a host machine and the AR.Drone.
All of our coding and control is processed on the host machine and
sent to the drone.

The AR.Drone has a multitude of sensors (accelerometers, gyroscopes,
sonar, and more) built in, which utilize for our learning purposes.

\section{Approach}

Given the large number of sensors and navigation data available on our
quadrotor platform, we suspected that there are unexploited
correlations between these sensor values and the drift of the
quadrotor.  It is difficult for a human to recognize any solid
correlation by looking at the data directly, but by using supervised
learning methods we have been able to discover relationships between
sensor values and quadrotor behavior, and use these to predict a large
portion of the drift.
 

To do this we recorded the onboard sensor data while simultaneously
collecting ground truth locations over time.  We collected the ground
truth values using the tracking system described in
\secref{trackingsystem}.  After applying the post-processing described
in \secref{data} to the collected data, we used supervised learning
methods to learn to predict the drone's drift.

For our supervised learning, we tested two support vector machine
(SVMs) libraries that support regression. The first algorithm we
tested was the LibSVM algorithm \cite{libsvm} in the Shogun machine
learning toolbox \cite{shogun}.  The second algorithm we tested is the
SVM-Light regression algorithm \cite{SVMLight} and obtained similar
results.  Our system only records and trains on drift in the
horizontal plane. We trained each axis (defined as X and Y) separately,
resulting in two separate trained systems.

\subsection{Tracking System}
\seclabel{trackingsystem} 

To collect ground truth data, we created a custom tracking system to
track MAVs in the horizontal plane.  This system uses an infrared LED
on the drone, and a Nintendo Wii remote (connected to the computer
though bluetooth) to track the MAV.  During data collection, we record
time, position from the Wii (in pixels), and 60 dimensions of data for
prediction.  This data comes from onboard sensors as well as the
navigation computer of the MAV.

To test our tracking system and to assist with data collection, we
constructed a simple PD controller to center the drone in real-time.
This system works well, and is demonstrated here:
\url{http://youtu.be/ptJ6E7jW2LY}.

\figvarp{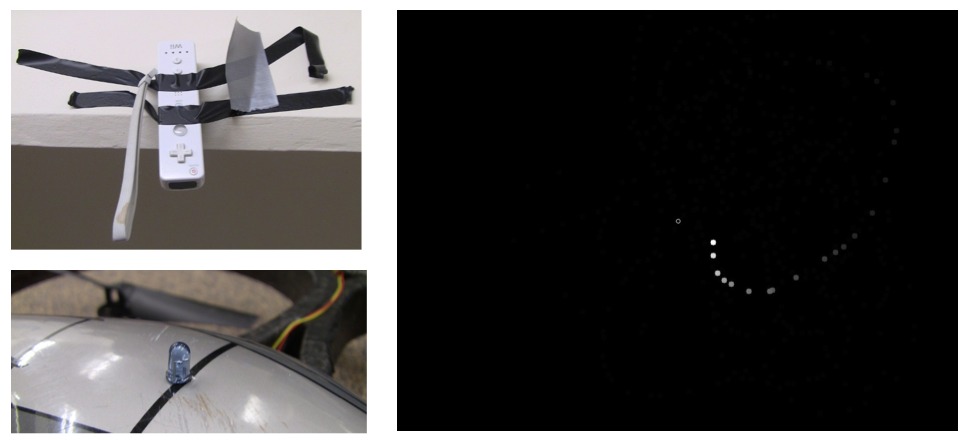}{1}{The tracking system developed to
  collect data for our project (\secref{trackingsystem}). \\Left: The tracking
  hardware\hspace{1em} Right: MAV being tracked.}{}

\subsection{Training Data}
\seclabel{data}

The training data we collected is divisible into three types:
\emph{time}, \emph{external data} available from the tracking system,
and \emph{onboard data} available from the quadrotor.  For training,
we learn a mapping from the onboard data to the derivatives --- pixels
per second in each of $x$ and $y$ --- computed using the time and the
external data from the wii.  During the testing or prediction phase,
we predict the derivatives from the onborad data.

The fields of data we collected, including all 60 dimensions of
onboard data, are shown below:

\begin{center}
\codebox{~~~~~~~Data fields collected and used for training and prediction\\
Time:~~~~~~~~~~~~~~~~~~~~time \\
Wii tracking system:~~~~~wii\_x, wii\_y, wii\_xd, wii\_yd, wii\_age,\\
\hspace*{0ex}~~~~~~~~~~~~~~~~~~~~~~~~~wii\_staleness \\
Onboard data:~~~~~~~~~~~~altitude roll pitch yaw vx vy vz acc\_x \\
\hspace*{0ex}~~~~~~~~~~~~~~~~~~~~~~~~~acc\_y acc\_z controlState vbat vphi\_trim \\
\hspace*{0ex}~~~~~~~~~~~~~~~~~~~~~~~~~vtheta\_trim vstate vmisc vdelta\_phi \\
\hspace*{0ex}~~~~~~~~~~~~~~~~~~~~~~~~~vdelta\_theta vdelta\_psi vbat\_raw ref\_theta \\
\hspace*{0ex}~~~~~~~~~~~~~~~~~~~~~~~~~ref\_phi ref\_theta\_I ref\_phi\_I ref\_pitch \\
\hspace*{0ex}~~~~~~~~~~~~~~~~~~~~~~~~~ref\_roll ref\_yaw ref\_psi rc\_ref\_pitch \\
\hspace*{0ex}~~~~~~~~~~~~~~~~~~~~~~~~~rc\_ref\_roll rc\_ref\_yaw rc\_ref\_gaz rc\_ref\_ag \\
\hspace*{0ex}~~~~~~~~~~~~~~~~~~~~~~~~~euler\_theta euler\_phi pwm\_motor1 pwm\_motor2 \\
\hspace*{0ex}~~~~~~~~~~~~~~~~~~~~~~~~~pwm\_motor3 pwm\_motor4 pwm\_sat\_motor1 \\
\hspace*{0ex}~~~~~~~~~~~~~~~~~~~~~~~~~pwm\_sat\_motor2 pwm\_sat\_motor3 \\
\hspace*{0ex}~~~~~~~~~~~~~~~~~~~~~~~~~pwm\_sat\_motor4 pwm\_u\_pitch pwm\_u\_roll \\
\hspace*{0ex}~~~~~~~~~~~~~~~~~~~~~~~~~pwm\_u\_yaw pwm\_yaw\_u\_I pwm\_u\_pitch\_planif \\
\hspace*{0ex}~~~~~~~~~~~~~~~~~~~~~~~~~pwm\_u\_roll\_planif pwm\_u\_yaw\_planif \\
\hspace*{0ex}~~~~~~~~~~~~~~~~~~~~~~~~~pwm\_current\_motor1 pwm\_current\_motor2 \\
\hspace*{0ex}~~~~~~~~~~~~~~~~~~~~~~~~~pwm\_current\_motor3 pwm\_current\_motor4 \\
\hspace*{0ex}~~~~~~~~~~~~~~~~~~~~~~~~~gyros\_offsetx gyros\_offsety gyros\_offsetz \\
\hspace*{0ex}~~~~~~~~~~~~~~~~~~~~~~~~~trim\_angular\_rates trim\_theta trim\_phi}
\end{center}

Over the course of this project, the following data points were
collected: 11,934 hovering, 1,640 with a directional command, and
417 with gusts of wind, for a total of 13,991 data points.  Plots
of a few fields of data over five capture sessions is given in
\figref{fiveruns}. 

\figvarp{fiveruns}{.75}{Some of the data collected from five manual
  runs.  The first plot is time, the next six are \emph{external} data
  from the Wii tracking system, and the rest represent a fraction of
  the 60 dimensions of \emph{internal} data from the quadrotor's
  sensors. Note that these graphs are not meant to be analyzed, but
  are presented merely as an example subset of the available features
  over time.}{}

Once the data is collected and logged to a file, we run
post-processing to make it more suitable for digestion by the learning
algorithms.  First, we remove all duplicate entries and \emph{stale}
entries caused by network delays, then we compute the position
derivatives (velocities) in pixels per second for both the $X$ and $Y$
directions, and it is this data that we learn on.
\figref{svd_not_norm} shows the singular values of the data before and
after normalization, to give an idea how many dimensions of
information are present in the data.

\figg{svd_not_norm}{svd_norm} {.48}{.48}{(left) Singular values of the
  60 dimensional training data before normalization. (right) Singular
  values of the 60 dimensional training data after normalization.  As
  one can see, the sensor data set is fairly rich, offering quite a
  few dimensions of non-covariant information.}

\section{Results}

By training on data collected in the lab, we were able to predict
quadrotor drift quite accurately.
\figref{dec17_1_DataLog-1292574599_07_x.pdf} shows the actual drift
velocity versus the drift velocity predicted by our learned model.
The predictions are not perfect, but they do match the actual drift
quite well, certainly much better than the null hypothesis of no drift
at all.

While \figref{dec17_1_DataLog-1292574599_07_x.pdf} shows results while
hovering and drifting in normal indoor conditions, in
\figref{dec17_windplus_40_DataLog-1292576304_00_x.pdf} we show results
obtained by creating an artificially windy environment.  We created
this environment by waving a large board back and forth, buffeting the
quadrotor.  A model was then trained on a combination of windy and
non-windy data and tested on other windy runs.  As one can see in the
figure, the quadrotor drifts significantly in one direction, and the
prediction follows quite well.

\figgg{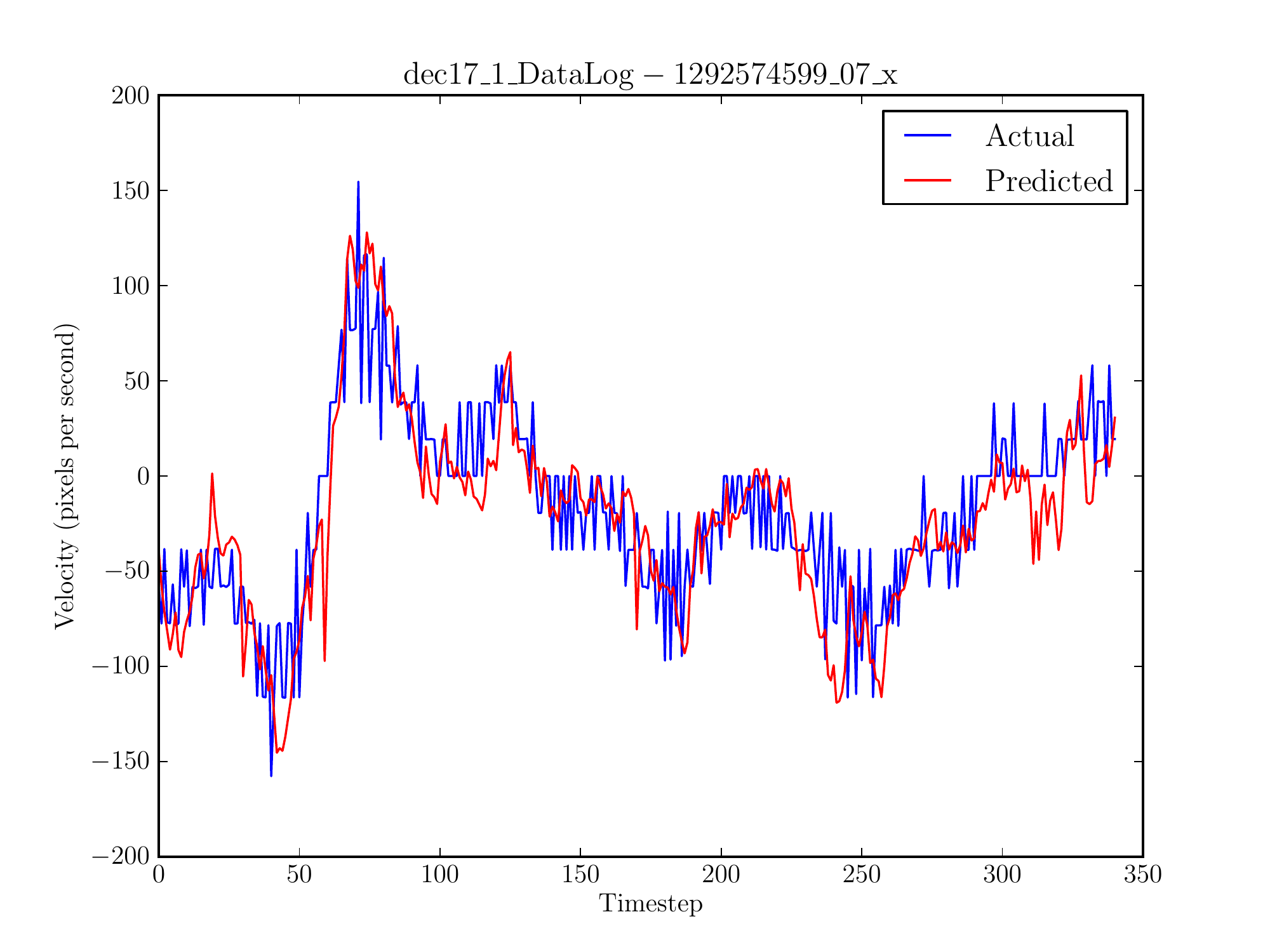}
      {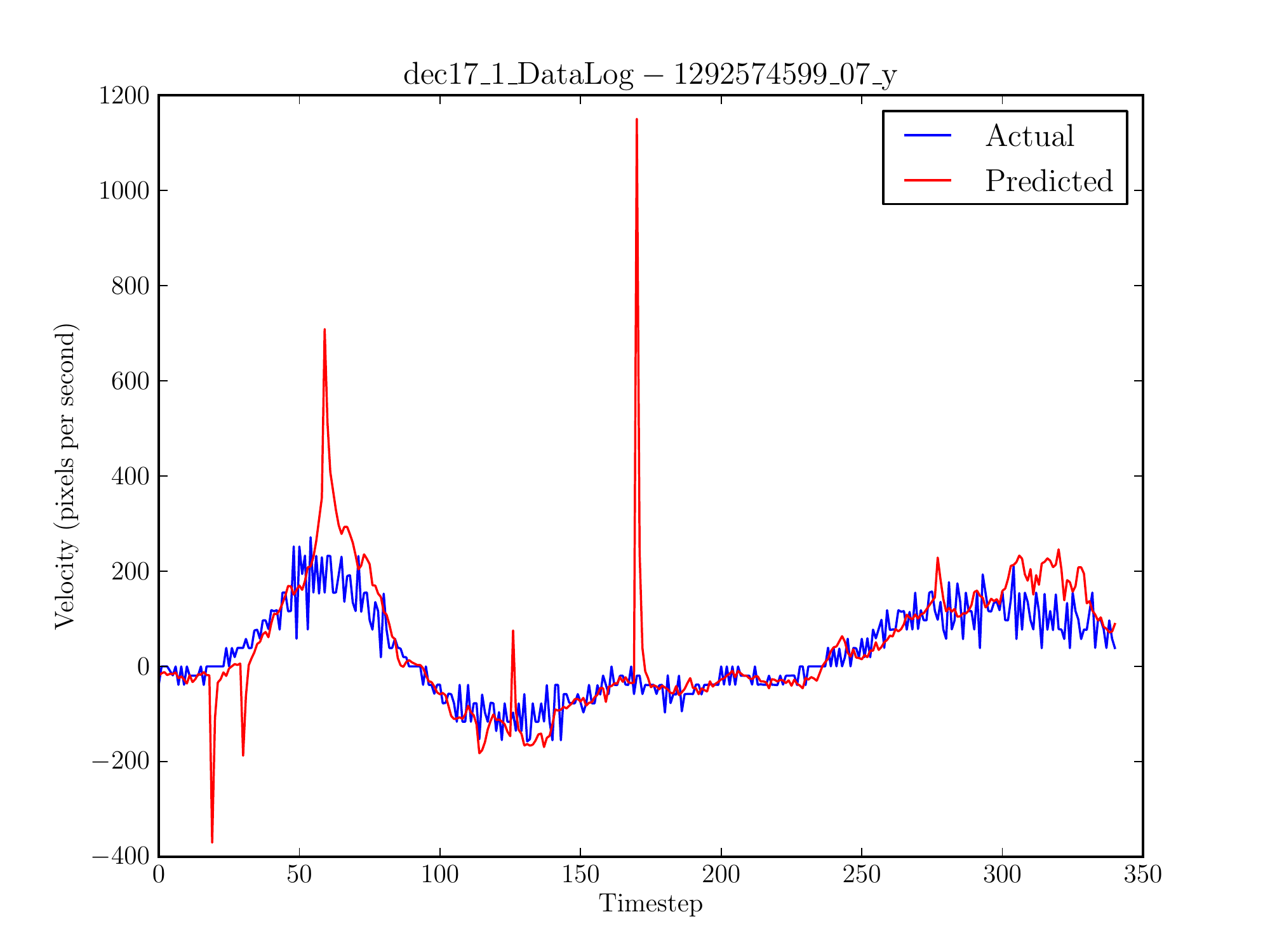}
      {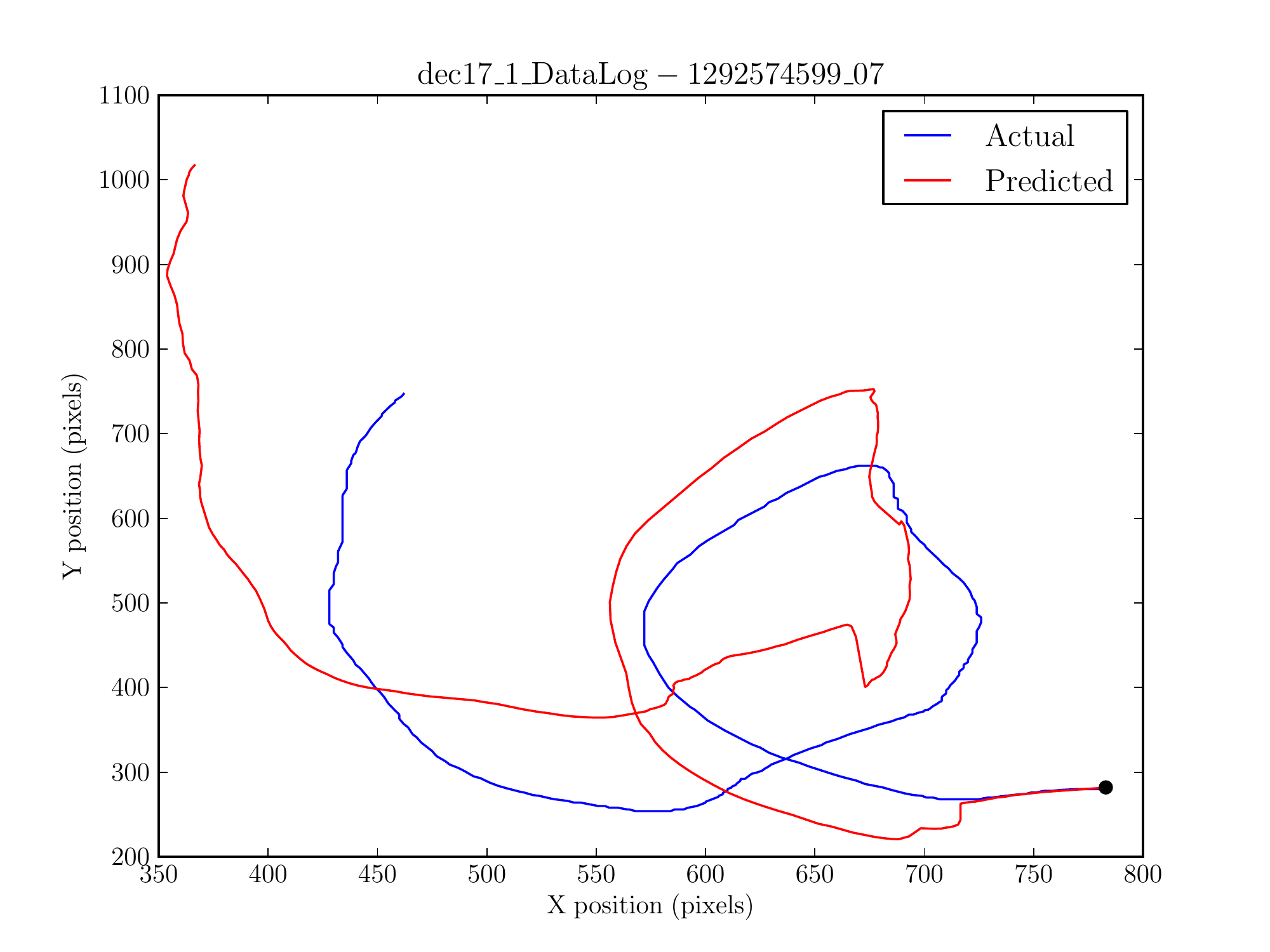} {.49}{.49}{1} {Results
        for normal indoor conditions.  Top left: predicted versus
        actual $x$ drift at each timestep in pixels per second. Top
        right: predicted versus actual $y$ drift at each timestep in
        pixels per second.  Bottom: actual position versus position
        inferred from integrating predictions over time.}

\figgg{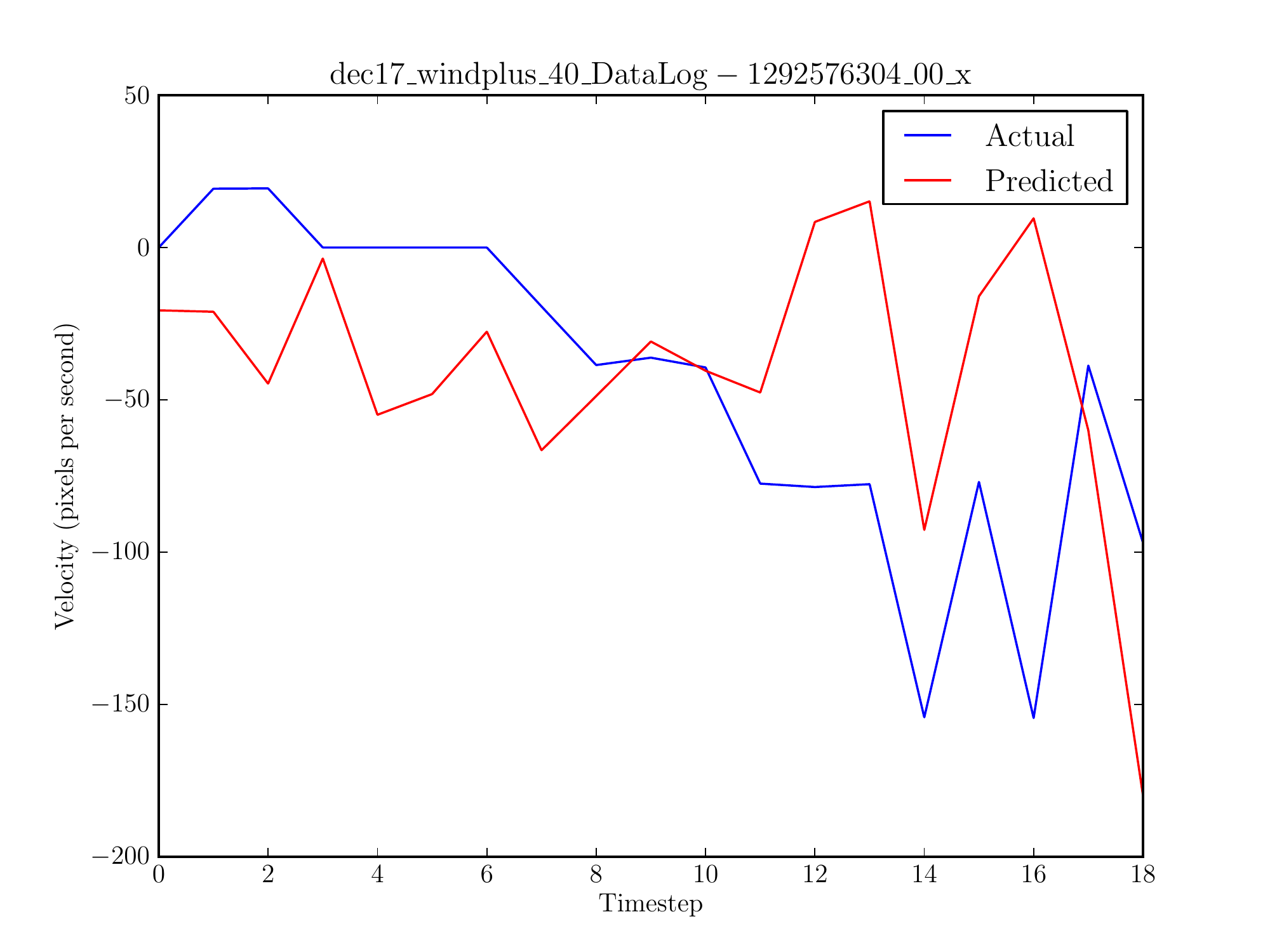}
      {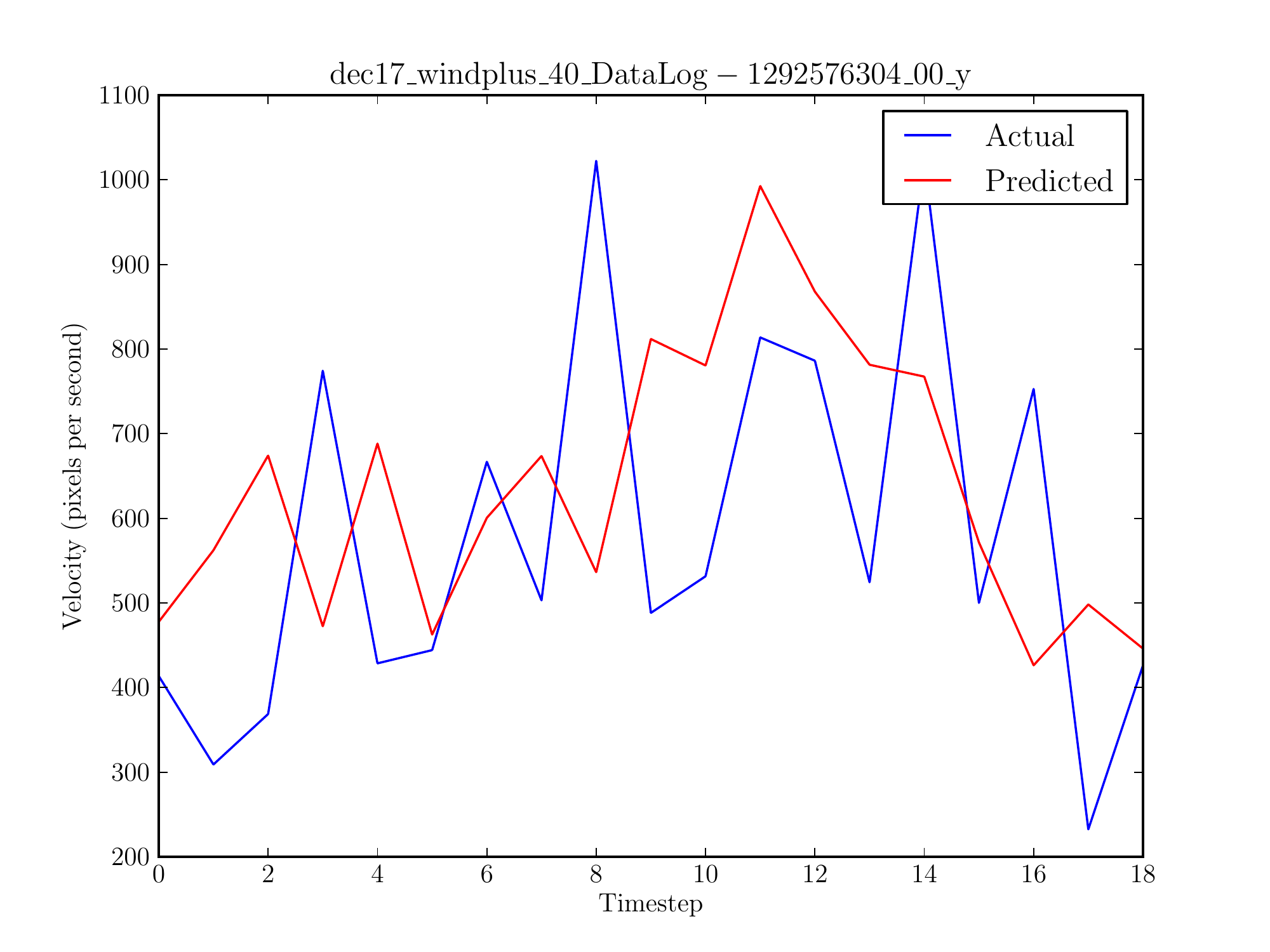}
      {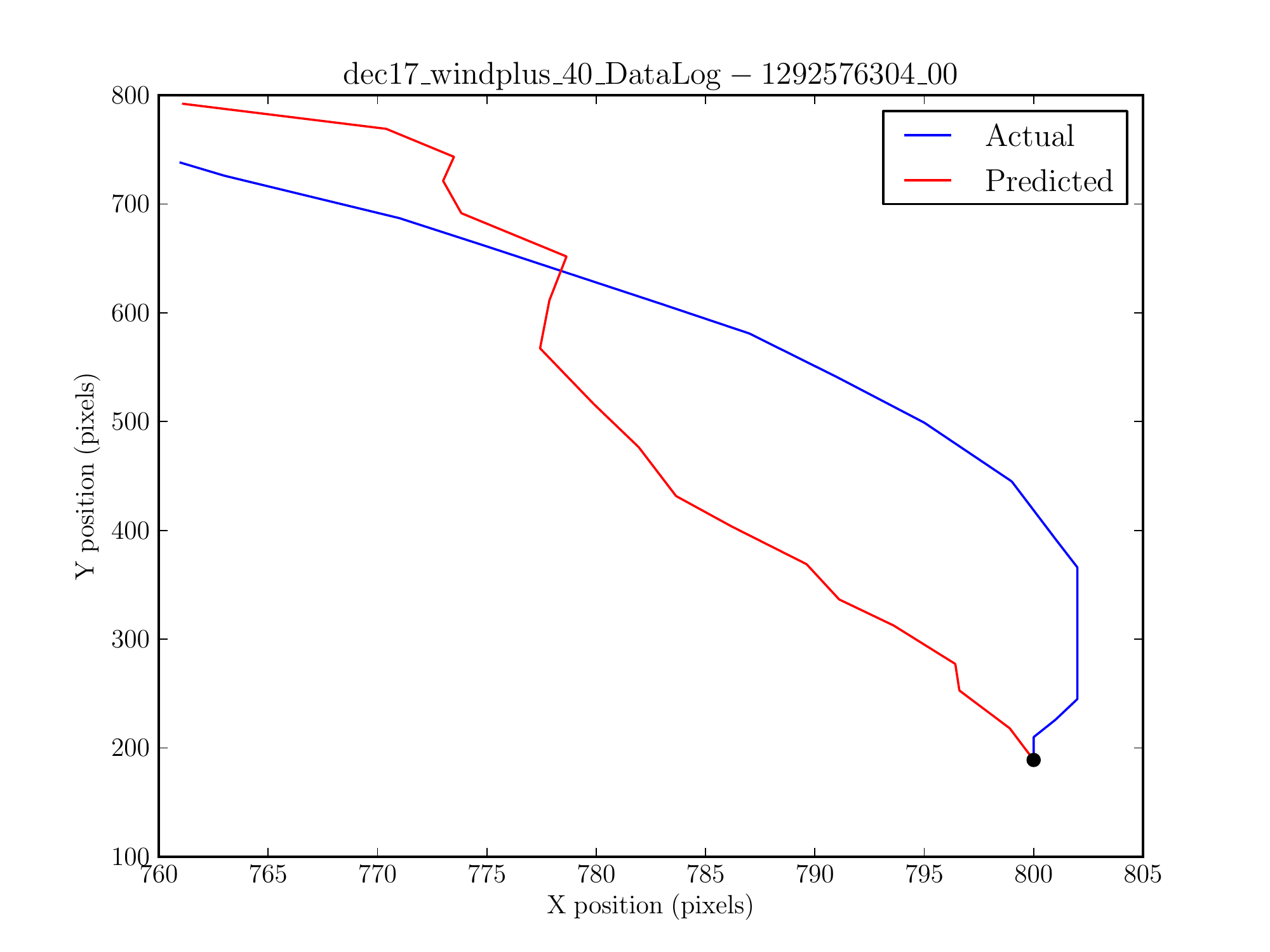} {.49}{.49}{1}
      {Results for environment with artificial wind.  This environment
        was created by waving a large board back and forth, blowing
        the quadrotor off to the side.  The learned model was trained
        on a combination of windy and non-windy data and then tested
        on other windy runs.  As one can see in the figure, the
        quadrotor drifts significantly in one direction, and the
        prediction follows quite well.  Top left: predicted vs. actual
        x drift.  Top right: predicted vs. actual y drift; Bottom:
        integrated position.}

\section{Future Work}

Although our algorithm was able to accurately predict drift, we are
still working on getting the prediction code to be fast enough to
control the quadrotor in live flight.  The SVM regression code runs
quickly, but we're currently facing difficulties with delays due to
I/O as well as a suboptimal configuration.  We integrated and tested a
complete feedback loop, but delays caused it to be unstable.

%
%

\section{Conclusion}

We are making strong progress towards predicting and removing the
quadrotor drift.  The drift is predicted fairly accurately via a 60
dimensional sensor data vector from the quadrotor, and we hope to soon
demonstrate a controller enhanced by the predicted drift.



\begin{thebibliography}{9}

\bibitem{Bills11} 
Cooper Bills, Yu-hin Joyce Chen and Ashutosh Saxena,
\emph{Autonomous Vision-based MAV Flight in Common Indoor Environments}, ICRA 2011. (Submitted)

\bibitem{shogun}
Soeren Sonnenburg, Gunnar Raetsch, Sebastian Henschel, Christian Widmer, Jonas Behr, Alexander Zien,
Fabio de Bona, Alexander Binder, Christian Gehl, and Vojtech Franc.
\emph{The SHOGUN Machine Learning Toolbox}. Journal of Machine Learning Research, 11:1799-1802, June 2010.

\bibitem{libsvm}
Chih-Chung Chang and Chih-Jen Lin, \emph{LIBSVM : a library for support vector machines}, 
2001. Software available at http://www.csie.ntu.edu.tw/~cjlin/libsvm

\bibitem{Nicol08}
Nicol, C.; Macnab, C.J.B. and Ramirez-Serrano, A. \emph{Robust neural network control 
of a quadrotor helicopter}, CCECE 2008.

\bibitem{Coza06}
Coza, C. and  Macnab, C.J.B. \emph{A New Robust Adaptive-Fuzzy Control Method Applied to
Quadrotor Helicopter Stabilization}, NAFIPS 2006. 

\bibitem{SVMLight}
T. Joachims, \emph{Making large-Scale SVM Learning Practical. Advances
 in Kernel Methods - Support Vector Learning}, B. Schölkopf and
C. Burges and A. Smola (ed.), MIT-Press, 1999.


\end{thebibliography}
\end{document}